\pdfoutput=1

\documentclass[11pt]{article}

\usepackage{ACL2023}

\usepackage{times}
\usepackage{latexsym}

\usepackage[T1]{fontenc}

\usepackage[utf8]{inputenc}

\usepackage{microtype}

\usepackage{inconsolata}

\usepackage{graphicx}
\usepackage{enumitem}
\usepackage{url}
\usepackage[textsize=scriptsize]{todonotes}

\usepackage{comment}

\title{Automatic and Human-AI Interactive Text Generation}

\author{Yao Dou\textsuperscript{\rm 1}\thanks{\hspace{4pt}Equal contribution.},~ Philippe Laban\textsuperscript{\rm 2}\footnotemark[1],~ Claire Gardent\textsuperscript{\rm 3}, Wei Xu\textsuperscript{\rm 1} \\ \textsuperscript{1} Georgia Institute of Technology \quad
  \textsuperscript{2} Salesforce \quad \textsuperscript{3} CNRS/LORIA \\ 
 {\small \tt{douy@gatech.edu} \quad
 \tt{plaban@salesforce.com} \quad
 \tt{claire.gardent@loria.fr} \quad
 \tt{wei.xu@cc.gatech.edu}}\\
}

\begin{document}
\maketitle

\paragraph{\textit{Note:}} \textit{This tutorial proposal has been accepted and we will give the tutorial at ACL 2024.}

\section{Description}

In this tutorial, we focus on text-to-text generation, a class of natural language generation (NLG) tasks, that takes a piece of text as input and then generates a revision that is improved according to some specific criteria (e.g., readability or linguistic styles), while largely retaining the original meaning and the length of the text. This includes many useful applications, such as text simplification \cite{Xu-EtAl:2015:TACL}, paraphrase generation \cite{narayan-etal-2017-split,iyyer-etal-2018-adversarial,goyal-durrett-2020-neural}, style transfer \cite{xu-etal-2012-paraphrasing, rao-tetreault-2018-dear, krishna-etal-2020-reformulating}, neutralizing biased language \cite{pryzant2019automatically,zhong-etal-2021-wikibias-detecting},  decontextualization for question answering \cite{DBLP:journals/corr/abs-2102-05169}, factual correction of human-written or machine-generated texts \cite{gao-etal-2023-rarr}, etc. 

In contrast to summarization \cite{liu-lapata-2019-text} and open-ended text completion (e.g., story) \cite{fan-etal-2018-hierarchical} where the length or content between input and output vary dramatically, the text-to-text generation tasks we discuss in this tutorial are more constrained in terms of the semantic consistency and targeted language styles. This level of control makes these tasks ideal testbeds for studying the ability of models to generate text that is both semantically adequate and stylistically appropriate. Moreover, these tasks are interesting from a technical standpoint, as they require complex combinations of lexical and syntactical transformations \cite{heineman2023dancing}, stylistic control \cite{kim-etal-2022-improving}, and adherence to factual knowledge \cite{devaraj-etal-2022-evaluating} --- all at once.

With a special focus on \textbf{text simplification and revision}, this tutorial aims to provide an overview of the state-of-the-art natural language generation research from four major aspects -- Data, Models, Human-AI Collaboration, and Evaluation -- and to discuss and showcase a few significant and recent advances: (1) the use of non-retrogressive approaches such as edit-based and diffusion models; (2) the shift from fine-tuning to prompting with large language models, which makes both human and automatic evaluation ever more important; (3) the development of new learnable metric and fine-grained human evaluation framework; (4) a growing body of studies and datasets on non-English languages; (5) the rise of HCI+NLP+Accessibility interdisciplinary research to create real-world writing assistant systems, bringing both excitements and ethical implications.

We will start by introducing the preliminaries for text simplification/revision, including the task formulation and high-quality datasets. We will present several representative neural and language models that excel in these tasks, and compare their strengths and weaknesses for generating different types of text edits and errors. We will highlight the newest developments in the Human-AI collaborative writing as well as in both human and automatic evaluations. We will conclude with discussions on ethical implications and future research directions. 

\vspace{3pt}
\noindent\textbf{Datasets For Text Revision Tasks.} Text Simplification is one of the more common text revision tasks and can serve as a good starting point for attendees of the tutorial. More specifically, we will introduce the Newsela corpus \cite{Xu-EtAl:2015:TACL}, a high-quality simplification resource in the news domain, and resources based on Simple Wikipedia, including Turk \cite{xu-etal-2016-optimizing}, ASSET \cite{fern2020asset}, CATS \cite{vstajner2018cats}, arXivEdits \cite{jiang-etal-2022-arxivedits}, and SWiPE \cite{laban-etal-2023-swipe}, gradually building from sentence-level to document-level datasets. We will then survey other text revision tasks, including tasks in the EditEval benchmark \cite{dwivedi2022editeval}, decontextualization \cite{choi-etal-2021-decontextualization}, and iterative codified editing \cite{du-etal-2022-understanding-iterative}. Finally, we will showcase available resources in different languages for text simplification, such as those in the MultiSim benchmark \cite{ryan-etal-2023-revisiting}, a growing collection of parallel corpora in different languages constructed collectively by researchers all over the world, which so far include: Arabic \cite{khallaf2022towards}, Basque \cite{10.1007/s10579-017-9407-6}, Brazilian Portuguese \cite{aluisio-gasperin-2010-fostering}, Danish \cite{klerke-sogaard-2012-dsim}, English, French \cite{gala-etal-2020-alector,grabar-cardon-2018-clear,cardon-grabar-2020-french}, German \cite{mallinson-etal-2020-zero,sauberli-etal-2020-benchmarking,aumiller-gertz-2022-klexikon,trienespatient,https://doi.org/10.48550/arxiv.1904.07733}, Italian \cite{miliani-etal-2022-neural,Tonelli2016,brunato-etal-2016-paccss,brunato-etal-2015-design}, Japanese \cite{maruyama-yamamoto-2018-simplified,katsuta-yamamoto-2018-crowdsourced,goto-etal-2015-japanese}, Russian \cite{Dmitrieva2021Quantitative,dmitrieva-tiedemann-2021-creating,sakhovskiy2021rusimplesenteval}, Slovene \cite{11356/1682}, Spanish \cite{orasan2013text,xu-etal-2015-problems,10.1145/2738046}, and Urdu \cite{qasmi-etal-2020-simplifyur}.

\vspace{3pt}
\noindent\textbf{Neural and Language Models for Generating Semantically Adequate Texts.} Various modeling approaches have been used for text revision and simplification.
We begin with edit-based models, a family well-suited to text revision given the substantial overlap
in content between inputs and outputs.
\citet{mallinson-etal-2020-felix} break down the task into sub-tasks of tagging tokens that are to kept and inserting missing tokens. \citet{mallinson-etal-2022-edit5} extend seq2seq models for an end-to-end manner. \citet{reid-neubig-2022-learning} propose multi-step editing setup, while
\citet{du-etal-2022-understanding-iterative} and \citet{kim-etal-2022-improving} incorporate edit intention and paragraph-level edits into the training process. \citet{agrawal-carpuat-2022-imitation} use a roll-in policy to generate realistic intermediate sequences and apply curriculum training.
Diffusion models \cite{ho2020denoising} is another family in non-autoregressive generation. Adapted to text generation by \citet{li2022diffusion}, these models are further tailored to conditional tasks such as text simplification and paraphrase generation \cite{gong2023diffuseq,yuan2022seqdiffuseq}.
We then discuss the use of autoregressive models for text revision. A simple and effective method is prepending of control tokens to the input during fine-tuning. These tokens can be sentence-level attributes \cite{martin2020controllable} or document-level edit plans \cite{cripwell-etal-2023-document}.
Iterative refinement with natural language instruction also gains attention, using either fine-tuned language models \cite{schick2022peer} or prompted LLMs\footnote{In the InstructGPT paper \cite{ouyang2022training} which preceded ChatGPT, OpenAI disclosed that ``Rewrite'' (text revision) made up 6.6\% of use cases in their API prompts, compared to 4.2\% for ``Summarization'' and 8.4\% for ``Chat''.} \cite{madaan2023self}.
Moreover, \citet{maddela-etal-2023-lens} guide text simplification with a metric using Minimum Bayes-Risk (MBR) decoding \cite{kumar2004minimum}.

\vspace{3pt}
\noindent\textbf{Human-AI Collaborative Writing.} 
The recent progress in LLMs abilities has led to many collaborative tools for writing, which involve realistic and diverse text revision tasks.
To provide a context on how human-AI collaboration are evaluated, we will first introduce some HCI academic work on co-writing tools prior to the existence of LLMs, such as Soylent \cite{Bernstein2010SoylentAW}, Wordcraft \cite{coenen2021wordcraft}, and MIL \cite{clark2018creative}.
Moving on to more recent developments, several work use fine-tuned LLMs for collaborative writing through iterative revision \cite{du-etal-2022-read} or interleaving planning and generation \cite{huot-etal-2023-text}, while \citet{lee2022coauthor} and \citet{jakesch2023co} use GPT-3 \cite{ouyang2022training} with prompting.
Finally, we will compare and contrast new commercial LLM-based text revision tools (such as Grammarly Go, QuillBot, ProWritingAid, etc.), and frame possible evaluation protocols that could fairly compare such systems (and future academic systems) based on the reviewed HCI work.

\vspace{3pt}
\noindent\textbf{Automatic and Human Evaluation.} Significant advancements have been made in both human and automatic evaluation methods for text revision over the past year. For automatic evaluation metrics, besides the commonly used SARI \cite{xu-etal-2016-optimizing} and BERTScore \cite{zhang2020bertscore}, \citet{maddela-etal-2023-lens} introduce the first learnable metric LENS that directly trained on human judgements. \citet{liu2023gpteval} use LLMs such as GPT-4 \cite{OpenAI2023GPT4TR} for evaluation.
Human evaluation can be categorized into intrinsic or extrinsic methods. In terms of intrinsic evaluation, most work prior to 2021 rely on 5-point Likert scale to evaluate various aspects of text quality, while \citet{alva-manchego-etal-2021-un} begin adopting direct assessment, rate from 0 to 100, for text simplification.
\citet{maddela-etal-2023-lens} incorporate ranking into direct assessment.
\citet{heineman2023dancing} propose a fine-grained, edit-level framework, which is further generalized into a platform for all types of fine-grained evaluations \cite{heineman2023thresh}.
On the extrinsic side, \citet{laban-etal-2021-keep} and others \cite{angrosh-etal-2014-lexico} evaluate text simplification through reading comprehension questions.

\section{Tutorial Outline} 

We will build in 10\%-15\% of time for Q\&A in each part of the tutorial.

\vspace{3pt}
\noindent\textbf{1. Introduction (15 min.)}

\begin{itemize}
[topsep=1pt,leftmargin=20pt,itemsep=-2pt]
  \item An overview of text generation research
  \item An overview of the tutorial
  \item Why text simplification and revision?
\end{itemize}

\vspace{3pt}
\noindent\textbf{2. Tasks and Datasets (30 min.)}
\begin{itemize}
[topsep=1pt,leftmargin=20pt,itemsep=-2pt]
    \item Text Simplification
    \item Other Text Revision Tasks
  \item Multilingual Resources
\end{itemize}

\vspace{3pt}
\noindent\textbf{3. Neural and Language Models (50 min.)}

\begin{itemize}
[topsep=1pt,=0,leftmargin=20pt,itemsep=-2pt]
    \item Edit-based models
    \item Diffusion models
    \item Conditional language models
    \item Large language Models with prompting
    \item MBR decoding
\end{itemize}

\vspace{3pt}
\noindent\textbf{4. Automatic and Human Evaluation (25 min.)}

\begin{itemize}
[topsep=1pt,leftmargin=20pt,itemsep=-2pt]
    \item Automatic Evaluation
    \item Human Evaluation
\end{itemize}

\vspace{3pt}
\noindent\textbf{5. Human-AI Collaborative Writing (20 min.)}

\begin{itemize}
[topsep=1pt,leftmargin=20pt,itemsep=-2pt]
    \item  Co-Writing tools Prior LLMs
    \item  Co-Writing tools in the era of LLMs
    \item  Commercial tools
\end{itemize}

\vspace{3pt}
\noindent\textbf{6. Live Demos: Text Revision Models and Human-AI Collaboration (10 min.)}

\paragraph{7. Ethical Considerations (20 min.)}

\paragraph{8. Conclusions and Future Directions (10 min.)}

\section{Tutorial Information}

\noindent\textbf{Type of the Tutorial.}  Cutting Edge. 

\vspace{3pt}
\noindent\textbf{Length.}  This is a 3-hour tutorial.

\vspace{3pt}
\noindent\textbf{Target Audience.} The tutorial will be accessible to anyone who has a basic knowledge of natural language processing. We think the topic will be of interest to researchers, students, and practitioners in both academia and industry. We will primarily target at the NLP community while bridging to a broader audience from HCI, Linguistics, Accessibility, Education, and other interdisciplinary fields. 

\vspace{3pt}
\noindent\textbf{Technical Equipment.} Standard AC setup and access to the Internet to show online demos. 

\vspace{3pt}
\noindent\textbf{Diversity Considerations.} We see a significant rise in the number of publications on the topic of Text Simplification in recent years, especially by researchers who work on non-English languages in many different countries. We will discuss resources and methods that support multilingual research, different text domains (e.g., medical and legal texts), and special user groups (e.g., deaf and hard-of-hearing users). The instructor team consists of researchers at different career stages (50\% female, 50\% male; 25\% Europe, 75\% U.S.): one senior researcher who is an ACL fellow, one professor, one Ph.D. student, and one industrial researcher. We will reach out to academic
communities such as WiNLP,\footnote{\url{https://www.winlp.org/}} SIGEDU,\footnote{\url{https://sig-edu.org/}} and Masakhane\footnote{\url{https://www.masakhane.io/}} to encourage participation from more diverse audiences.

\vspace{3pt}
\noindent\textbf{Estimated Size of Audience.} Given the rising interests in text generation (text simplification and revision in particular), we estimate the tutorial will attract an audience of 100+, based on the recent success of Workshop on Text Simplification, Accessibility, and Readability (TSAR) at EMNLP 2022 and the birds-of-a-feather event (over 50 participants) on Text Simplification at NAACL 2021.

\vspace{3pt}
\noindent\textbf{Venues.} We prefer ACL and NAACL due to travel constraints of some speakers. EMNLP is our third option, and we currently do not consider EACL.

\vspace{3pt}
\noindent\textbf{Breadth of Research Covered. } This tutorial is intended to cover recent research advances from both academia and industry. We estimate that 20\% of the work covered in this tutorial will be from the presenters and the remaining 80\% by others. 

\vspace{3pt}
\noindent\textbf{Open Access.} We will build a website for the tutorial and share all teaching material freely online. We also agree to allow the publication of slides and video recordings by ACL.

\section{Tutorial Instructors} 

\noindent \textbf{Yao Dou} (\url{https://yao-dou.github.io/}) is a Ph.D. student in the College of Computing at the Georgia Institute of Technology, advised by Prof. Wei Xu. His research interests lie in natural language processing and machine learning. His recent work focuses on text simplification, text evaluation, and social media privacy.

\vspace{3pt}
\noindent \textbf{Philippe Laban} (\url{https://tingofurro.github.io/}) is a Research Scientist at Salesforce Research. His research is at the intersection of NLP and HCI, focusing on several tasks within text generation, including text simplification and summarization. He received his Ph.D. in Computer Science from UC Berkeley in 2021. His thesis is titled ``Unsupervised Text Generation and its Application to News Interfaces''. His recent work has focused on expanding the scope of text simplification to the paragraph and document-level and evaluating text-editing interfaces. He publishes in both *ACL and HCI conferences, including work on interactive user interface design for NLP applications. 

\vspace{3pt}
\noindent \textbf{Claire Gardent} (\url{https://members.loria.fr/CGardent/}) is a senior research scientist at the French National Center for Scientific Research (CNRS), based at the LORIA Computer Science research unit in Nancy, France. In 2022, she was selected as an ACL Fellow and was awarded the CNRS Silver Medal. She works in the field of NLP with a particular interest in Natural Language Generation. In 2017, she launched the WebNLG challenge, a shared task where the goal is to generate text from Knowledge Base fragments. She has proposed neural models for simplification and summarization; for the generation of long-form documents such as multi-document summaries and Wikipedia articles; for multilingual generation from Abstract Meaning Representations and for response generation in dialog. She currently heads the AI XNLG Chair on multi-lingual, multi-source NLG and the CNRS LIFT Research Network on Computational, Formal and Field Linguistics. 

\vspace{3pt}
\noindent \textbf{Wei Xu} (\url{https://cocoxu.github.io/}) is an Associate Professor in the College of Computing at the Georgia Institute of Technology. Her recent research focuses on text generation (including data construction, controllable model, human and automatic evaluation), stylistics, analyzing and evaluating large language models (including multilingual capability, cross-lingual transfer learning, cultural bias, and cost efficiency). She is a recipient of the NSF CAREER Award, CrowdFlower AI for Everyone Award, best paper award from COLING 2018, and honorable mention from ACL 2023. She is an NAACL executive board member and regularly serves as a (senior) area chair for *ACL conferences. She frequently gives invited talks at universities and companies. She has given tutorials on ``NLP for Social Media and Text Analysis'' and has organized multiple workshops, including WNUT, GEM, and TSAR. 

\section{Ethical Statement}

It is important to consider the ethical implications of text simplification and revision research, especially in the realizations of them as assistive technologies \cite{gooding-2022-ethical}. In the tutorial, we will cover many ethical considerations of these technologies, including the distinctions and varied needs of different targeted user groups (e.g., non-native speakers, children, and people with dyslexia), best practices of crowdsourcing and user studies, privacy preservation and safety concerns (e.g., when simplifying medical texts), as well as relevant research findings in \cite{10.1145/3520198,10.1145/3589955}.

\section{Reading list}

\begin{itemize}
[topsep=2pt,leftmargin=*,itemsep=-2pt]
  \item SWiPE: A Dataset for Document-Level Simplification of Wikipedia Pages \cite{laban-etal-2023-swipe}
  \item LENS: A Learnable Evaluation Metric for Text Simplification \cite{maddela-etal-2023-lens} 
  \item RewriteLM: An Instruction-Tuned Large Language Model for Text Rewriting \cite{shu2023rewritelm}
  \item PEER: A Collaborative Language Model \cite{schick2023peer}
  \item DiffuSeq: Sequence to Sequence Text Generation with Diffusion Models \cite{gong2023diffuseq}
  \item MUSS: Multilingual Unsupervised Sentence Simplification by Mining Paraphrases \cite{martin-etal-2022-muss}
  \item On the Ethical Considerations of Text Simplification \cite{gooding-2022-ethical}
  \item Making Science Simple: Corpora for the Lay Summarisation of Scientific Literature \cite{goldsack-etal-2022-making}
  \item Findings of the TSAR-2022 Shared Task on Multilingual Lexical Simplification \cite{saggion-etal-2022-findings}
  \item Paragraph-level Simplification of Medical Texts \cite{devaraj-etal-2021-paragraph}
  \item Neural CRF Model for Sentence Alignment in Text Simplification \cite{jiang-etal-2020-neural}
   \item CoAuthor: Designing a Human-AI Collaborative Writing Dataset for Exploring Language Model Capabilities \cite{lee2022coauthor}
   \item Thresh: A Unified, Customizable and Deployable Platform for Fine-Grained Text Evaluation \cite{heineman2023thresh}

\end{itemize}

\section*{Acknowledgements}
The presenters' research is supported in part by the NSF awards IIS-2144493 and IIS-2112633, ODNI and IARPA via the HIATUS program (contract 2022-22072200004). The views and conclusions contained herein are those of the authors and should not be interpreted as necessarily representing the official policies, either expressed or implied, of NSF, ODNI, IARPA, or the U.S. Government. The U.S. Government is authorized to reproduce and distribute reprints for governmental purposes notwithstanding any copyright annotation therein.

\bibliography{anthology,custom}
\bibliographystyle{acl_natbib}

\end{document}